# A Marketplace Price Anomaly Detection System at Scale

Akshit Sarpal
Walmart Global Tech
akshit.sharpal@walmart.com

Qiwen Kang
Walmart Global Tech
qiwen.kang@walmart.com

Fangping Huang
Walmart Global Tech

fangping.huang@walmart.com

Yang Song
Walmart Global Tech
yang.song@walmart.com

Lijie Wan
Walmart Global Tech
lijie.wan@walmart.com

## ABSTRACT

Online marketplaces execute large volume of price updates that are initiated by individual marketplace sellers each day on the platform. This price democratization comes with increasing challenges with data quality. Lack of centralized guardrails that are available for a traditional online retailer causes a higher likelihood for inaccurate prices to get published on the website, leading to poor customer experience and potential for revenue loss. We present *MoatPlus* (**M**asked **O**ptimal **A**nchors using **T**rees, **P**roximity-based **L**abeling and **U**nsupervised **S**tatistical-features), a scalable price anomaly detection framework for a growing marketplace platform. The goal is to leverage proximity and historical price trends from unsupervised statistical features to generate an upper price bound. We build an ensemble of models to detect irregularities in price-based features, exclude irregular features and use optimized weighting scheme to build a reliable price bound in real-time pricing pipeline. We observed that our approach improves precise anchor coverage by up to 46.6% in high-vulnerability item subsets.

## 1 INTRODUCTION

Major non-luxury retailers have maintained a significant pricing discipline as consumer decision to make a purchase is often heavily driven by the product price. Walmart is the largest retailer offering a hybrid fulfillment model with three broad channels - physical stores, first party catalog where the items are purchased, priced, and fulfilled by Walmart and a third-party (marketplace) platform where pricing is controlled by an independent seller. For first-party items, our team built a price anomaly detection system based on a relatively rich feature space [5], however there were multiple data challenges that made its application to a marketplace setting infeasible. Their work leverages features such as (1) *cost* – which can set a reliable guardrail for item price, (2) *inventory* – which can explain if the price is adjusted upwards due to low item availability and (3) *hierarchy* – which can group similar so their commonalities can be exploited for setting price guardrails. Above features are either not available with the marketplace platform or are unreliable being ingested from sellers.

In the current state, Walmart marketplace pricing pipeline uses independent features called **anchor prices** along with analytics-based multipliers to create an upper price bound called **ceiling price**. Some examples of item-level anchor prices include prices from various competitor websites, MSRP, Walmart store price and Walmart first-party listing price. Each item may have one or more anchor prices available in the pricing pipeline, and we select one of the available anchor prices using proprietary business heuristics. The selected anchor price is then used to build a ceiling price, and if a marketplace offer price breaches the ceiling price, it gets unpublished. The primary challenge with this approach is that the individual anchor prices are also prone to be anomalous which can lead to anomalously low or high ceiling prices.

To ensure high customer and seller experience on our marketplace platform, ceiling prices need to be precise and reliable. An egregiously high ceiling price can lead to anomalously high-priced offers being visible on the website causing negative customer price perception, which for a value-based retailer like Walmart could hurt its core EDLP (Every Day Low Price) principle. On the other hand, a low ceiling price can lead to falsely delisting a reasonably priced item eroding marketplace seller trust. Egregiously high prices not only impact customer price perception of Walmart, but also result in lost revenue due to forgone sales and are therefore imperative to detect. We focus on detecting high price anomalies in this paper.

One approach to enhancing the existing heuristics-driven anomaly detection system is to leverage historical values of multiple anchor prices to generate a single item-level **optimal anchor price** (and therefore ceiling price) in an offline setting. Given the dynamic nature of item prices (due to promotional events, seasonality, etc.), we observed that offline and static item-level ceiling prices pose two challenges: (1) *relevance* – ceilings get stale quickly and lead to increased misclassifications and (2) *cold start* – newly set-up items are not covered due to unavailability of historical anchor price data at time of offline ceiling price generation. To address these, we built an online system that detects anomalous anchor prices and uses the remaining anchor prices to build an optimal anchor price, which is an estimate of reasonable item price. This optimal anchor price is used to build a ceiling price that serves as item-level upper price bound.

We apply weak supervision for bootstrapping labeled data and implement a feedback loop to generate additional labels using internal tools and operations team. While the primary objective is to build a ceiling price using anchor prices for the purpose of blocking egregious offer prices on Walmart.com, our underlying models also help flag issues with individual anchor prices, which can be used as feedback for correcting errors from anchor-generation sources. Moreover, our system design provides a highly modular approach that allows us to integrate with additional anchor price sources as they become available in future (for example price stream from a new competitor). While our work focusses on detection of anomalous prices in a marketplace platform setting, we believe that our learnings can be applicable for anomaly detection for other use-cases where multiple independent data sources estimate some common phenomenon, such as data from multiple sensors. The novel contributions for this work, compared with previously published work are as follows:

- **Highly scalable anomaly detection framework for marketplace platform pricing** – while there are applications of pricing anomaly detection to ecommerce pricing [5], they do not particularly address challenges specific to online marketplaces where limited and lower quality data is available due to a decentralized item setup process; our work focusses on online marketplaces where other rich features may not be reliably available and price update volumes are over 25 times higher.

- **Highly modular and robust detection approach for aberrant anchors** – our approach of fitting separate detector models for individual anchor prices ensures high flexibility to add new detector models and high coverage when comparing with a consolidated singular model.

- **An application of weak-supervision to price anomaly detection** – labeled data acquisition and denoising requires significant time and resources to set-up and maintain; we leverage weak supervision [4,9,11] to label data and reduce significant overhead of engineering and crowdsourcing resources, allowing us to rapidly build individual supervised anomaly detection models for a subset of anchor prices.

- **Optimized weight selection and price bound generation** – an **interpretable** solution to anomaly detection is imperative to maintain seller trust while delisting egregious offers. We hypothesize that under different conditions, certain anchor prices tend to be more reliable than others. For example, it may be more complex to match items between Walmart and competitor catalog in *T-shirt* category than *Books* category due to absence of hard identifiers like ISBN numbers. This may lead to a higher mismatch rate and pose a risk that we use an incorrect competitor item price as an anchor price for the given Walmart item. Consequently, we may end up with an incorrect ceiling price, even though other reliable anchor prices were available. To overcome this challenge, we train a multiclass classification model to predict which anchor prices tend to be more reliable given certain item characteristics so they can be assigned a higher weight. Using classifier probabilities, we take a weighted mean of available anchor prices to build a consolidated and interpretable optimal anchor price.

The *MoatPlus* system strikes a balance between accuracy, latency, and interpretability. The remainder of this paper is organized as follows: we start with a review of common anomaly detection methodologies and use-cases in section 2, followed by the description of our data, features, methodology and design in section 3. We set up various experiments and report on the experimental performance in section 4. Finally, we conclude with the system performance in production setting.

## 2. LITERATURE REVIEW

Anomaly detection is a widely researched domain with diverse methodologies proven to work for multitude of use-cases. Detection techniques can be broadly categorized into (1) Statistical based [19], which leverage the assumption that normal data occur in high-probability regions of a stochastic model while anomalous instances occur in low-probability regions; (2) classification based [10], where a classifier is built to distinguish between normal and anomalous classes and can include models like neural network, SVMs, Bayesian networks, or tree-based models; (3) Nearest Neighbor based [20,21] that analyze each instance with respect to its local neighborhood, and can include methods like k-NN and Local Outlier Factor (LOF) [3]; (4) Clustering based [20] methods that assume that the normal data instances lie within clusters or closer to centroids and anomalous instances do not. Classification-based methods are generally faster than other approaches at inference-time but rely on availability of accurate labels. We implement weak supervision for data labeling and evaluate various classification-based models this paper.

Weak supervision is an approach to machine learning in which high-level and often noisier sources of supervision are used to create much larger training sets much more quickly than could otherwise be produced by manual supervision [20]. Common sources of noisy labels include heuristic patterns, crowd labels, external knowledge bases (distant supervision) [11] and feature annotations [12]. We focus on an implementation involving heuristic patterns.

Statistical anomaly detection methods can be further classified into (1) parametric techniques which assume the knowledge of underlying distribution and estimate the parameters from the given data [14] and (2) non-parametric techniques that do not generally assume knowledge of underlying distribution [13]. Some applications of anomaly detection demonstrate that mapping the data to a feature space based on kernel density estimation provide significant lift in classification performance [26]. We leverage this idea in our paper and generate a density estimate based feature to encode local information for classifier-based anomaly detection.

For optimized weighting of non-anomalous anchor prices, we use multiclass classification, where we either get majority vote from an ensemble of base learners or get average probability of belonging to a class using an ensemble of probabilistic classifiers. Some classifiers provide a natural way to extend the binary classification task to multiclass scenarios while others are inherently binary. Some common approaches for adapting an inherently binary classifier to multiclass use-case like *one-vs-one* and *one-vs-rest* are available in literature. We explore both types of classifiers in this paper.

Anomaly detection has been applied to a variety of use-cases including fraud detection, detecting medical anomalies and disease outbreaks, industrial damage detection and sensor networks [1,2]. More modern use-cases include efficiency regression detection in modern software development cycle [16], extreme event forecasting [17], and cyber-intrusion and abuse detection [18]. We extend the applications for ecommerce pricing use-case.

## 3. METHODOLOGY

In this section, we describe the new architecture design of Walmart marketplace high-price anomaly detection system. We first provide an overview of the system design. Next, we cover the data and features used to train the underlying models, introduce the three layers that make up the system. Lastly, we summarize the evaluation metrics that measure performance of the system.

### 3.1 System Design

We provide a high-level overview of the system built on the components. The marketplace pricing pipeline receives ~100MN pricing or anchor update events a day. A majority of these events trigger a ceiling price recalculation in real-time. A **ceiling price** is defined as an upper price bound for an item – if the incoming price from marketplace seller is higher than this bound, it is deemed as egregiously high. The calculated real-time ceiling price for each item is passed downstream to a tier zero system which assigns it to each offer belonging to that item (an item can have multiple offers due to multiple sellers selling a common item).

When an offer price is deemed anomalous, the offer gets removed from the website and the corresponding seller gets notified of the delisting through a centralized internal application. The seller can review the notification and either update the price or dispute the delisting if they believe the price is reasonable. If they update the price, the new price update event gets passed to the pipeline again which triggers a recalculation and gets republished if the updated price is under the ceiling price. If the seller disputes the price block and the offer price is verified as reasonable, it indicates the ceiling was too low. A manual override is applied to the ceiling, and all offers that are now below the overridden ceiling are republished. Figure 2 provides a high-level summary of the system.

For implementation, we train and store the models as objects in a filesystem. The models are retrained on-demand when either additional labeled data becomes available, or we observe performance drift. We package *MoatPlus* as a python library and distribute it via internal PyPi server. The library is deployed as an API hosted on various compute nodes. There are three layers in the system – KDE feature generation layer, detector layer and aggregator layer. Each call to the *MoatPlus* API goes to a centralized model interface, which orchestrates the layers. In response, three values are returned – (1) an **optimal anchor**, (2) a dictionary of individual anchor weights corresponding to each anchor for calculating the optimal anchor, and (3) Boolean indicators to report whether any anchor values were flagged as anomalous by the detector layer models. Optimal anchor is considered as an estimate of reasonable item price. It is multiplied by a business-provided constant to build a ceiling price. This constant serves as a tolerance buffer denoting how far an offer price needs to deviate from reasonable price to be classified as anomalous. Figure 1 shows the high-level overview of the *MoatPlus* library.

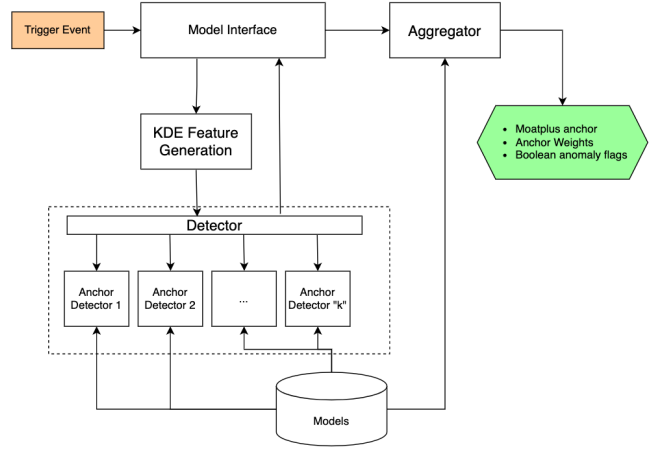

**Figure 1.** *MoatPlus* **library design – a high-level overview.**

In total, we train four classification models offline – three for the detector layer and one for the aggregator layer. These models are invoked for inference in the real-time pricing pipeline. The feature space, data preparation and training procedure for the underlying models is demonstrated in the remainder of the section.

### 3.2 Features

We summarize the features we used for the underlying models within the system in this section. We represent the feature vector by $x$, where $x_i$ represents the $i^{th}$ feature. Each of the underlying models uses different features. We aim to bucket the features into broad segments for explainability.

**Real-time price anchors.** A price anchor is a feature resembling a price point. We currently leverage five anchor prices available in the system, and the modular nature of our design can enable seamless integration with additional anchor prices in future. In this paper, they are denoted by $A$ and if $x_i \in A$, then it is a price anchor.

**Markup-based features.** We first define ratio-based features $R_i$ for real-time anchor prices $x_i \in A$ using median of observed marketplace prices $x_0$ as a base. These features provide a notion of the relative distance of an anchor price from the median marketplace price.

$$R_i = \ln \frac{x_i + c}{x_0 + c} \ \forall \ x_i \in A \ . \tag{1}$$

These log-ratios are not identically distributed because certain anchors tend to be higher than others in general; for instance, competitors' ecommerce prices tend to be higher than respective item's store prices due to the overhead of fulfillment costs. On the other hand, a higher competitor price does not make it more likely to be anomalous. Therefore, we further transform $R$ to $M$ using the below transformation. This transform ensures $x_i \in M$ are identically distributed,

$$M_i = \frac{R_i - m_i}{s_i} \ \forall \ i \in \{1,..,k\}, \qquad (2)$$

where $m_i$ and $s_i$ represent pre-learned estimates of population means and standard deviations for the $i^{th}$ ratio-based feature, $R_i$. Markup-based features provide the notion of relative distance of an anchor from an observed price with respect to the "typical" distance of that anchor from observed marketplace price. Applying the transformations has two advantages – we achieve identical distribution to a reasonable degree and features are condensed to a smaller scale which makes it easier to generalize bandwidth selection for kernel density estimation. Figure 3 shows an example of such two anchors.

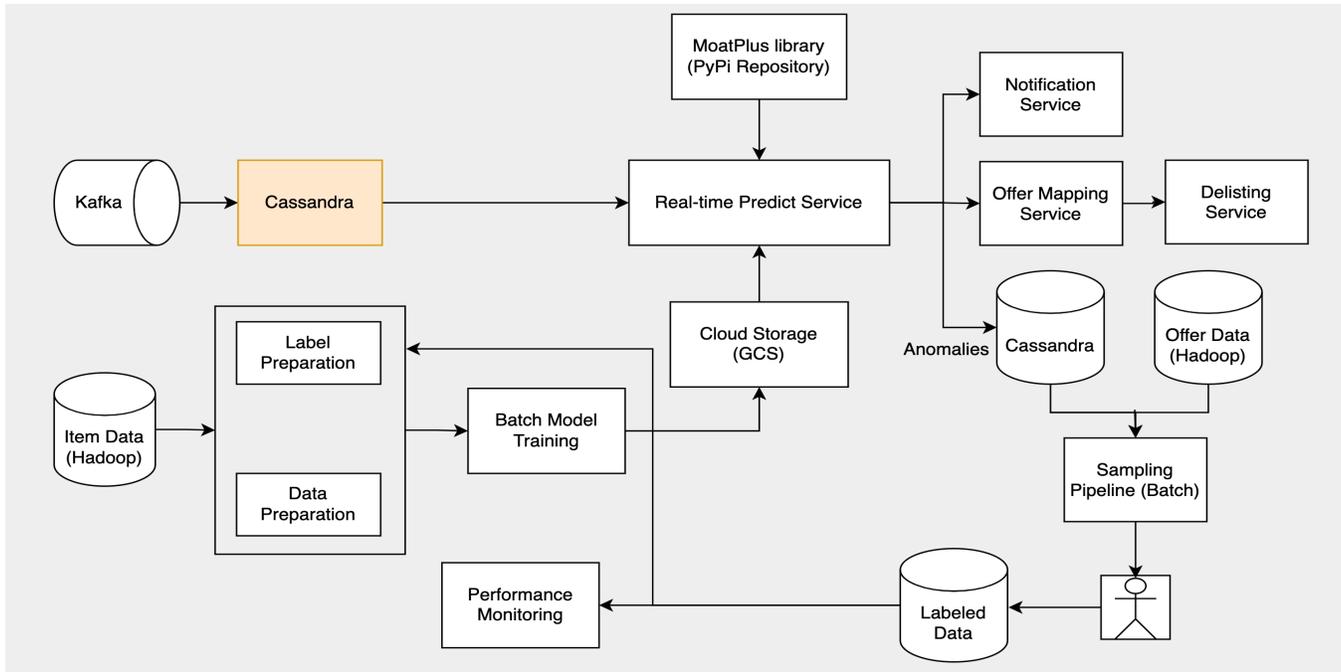

**Figure 2. System Architecture.**

**Density-based transformations**. We leverage kernel density estimation to fit a distribution for markup-based transformations $M$, and use density estimates to represent a notion of closeness between real-time anchors $A$. Markup-based values occurring in dense neighborhoods will have higher density scores. We observe that these estimates learn the scale-adjusted similarity connotations between anchors, which helps the models in detector layer to isolate anomalous anchors more effectively. Density-based features are denoted as $D$ in our work. Therefore, we learn mappings from $M$ to $D$ such that the mapping $y_i \in D$ is higher if $x_i \in M$ occurs in a close neighborhood of $x_j \in M$ for $j \in \{1,..,k\}$ and $j \neq i$.

**History-based features**. Historical price anchor ranges can be extremely informative to assess whether a given anchor is aberrant.

However, we observe that the historical anchor price data is contaminated for a meaningful segment of the catalog. There are items where majority of price anchor data is anomalously high. We leverage a combination of unsupervised learning and rule-based logic to cleanse historical anchor data and create statistical features that are used by the detector layer. These historical outlier-cleansed statistical features are denoted as $H$.

**Competing marketplace offer prices**. Information on other sellers' price for the same offer is an important guide for assessing price quality. We take real-time offer prices from other sellers as an array and build statistical features like mean and coefficient of variation. These are denoted by $O$ in our work.

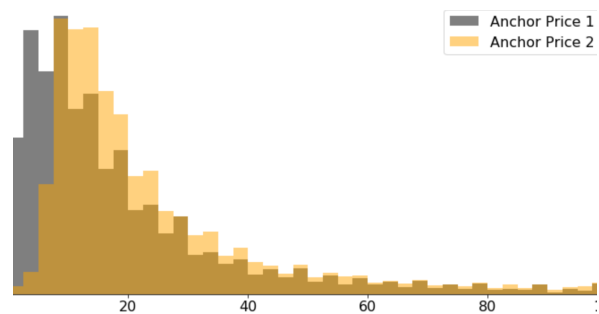

**Figure 3. Distribution of two anchor prices for a sample of 10,000 items. Anchor Price 2 is shifted right indicating that it tends to be higher than Anchor Price 1. Upon rescaling, both anchor prices can be used for estimating true reasonable price.**

**Table 1**. Summary of feature sets.

| Feature-set | Symbol | Examples |
|---|---|---|
| Real-time price anchors | $A$ | CompetitorPrice |
| Markup-based features | $M$ | Log (1pPrice / MpPrice) |
| Similarity-based transformations | $D$ | KD Estimates of $M$ |
| History-based features | $H$ | MinPastCompetitorPrice |
| Competing offer features | $O$ | Range (MpOfferPrice) |

### 3.3 KDE Feature Generation Layer

Nonparametric methods for probability density estimation assume no functional form of the distribution and aim to learn the distribution from observed samples. For a given item, we take elements of Markup-based feature vector $x_i \in M$ as *i.i.d.* samples from an arbitrary distribution. Kernel Density Estimation (KDE) (also known as Parzen-Rosenblatt window method) is a widely used non-parametric approach to estimate a probability density function $p(\mathbf{x})$ for a specific point, $\mathbf{x}$ from a sample $\{\mathbf{x}_n\}$ that doesn't require any knowledge or assumption about the underlying distribution. The probability density estimate is given as

$$p(x) = \frac{1}{kh}\sum_{i=1}^{k} K(\frac{x - x_i}{h}) \qquad (3)$$

where $k$ represents number of samples, $K$ denotes a kernel function and $h$ denotes kernel bandwidth. There are two tunable hyperparameters for estimation – kernel and bandwidth. A common approach to tuning these parameters is through cross-validation [6,7,8]. Alternately, it is possible to optimize the bandwidth parameter directly by incorporating in the classifier's objective function [15]. For a real-time application of KDE, we explored simpler alternatives since tuning bandwidth at inference-time introduces additional latency. We shortlisted two practical approaches to bandwidth-selection (1) *rule-of-thumb* approaches available in literature [7] like Silverman's rule and Scott's rule [22] and (2) a *trial-based* approach where we shortlisted few values and performed trials. To validate our choice of bandwidth we built a set of robust test-cases on the expected behavior of estimates and their separation properties. We found that a Gaussian kernel with a bandwidth of 0.5 outperforms *rule-of-thumb* approaches on the test cases.

It is important to note that not all items have all $k$ anchor prices available. In fact, anchor prices can be fairly sparse for some items with as little as just one anchor price available. In absence of additional information, items that have only single anchor price available would demonstrate highest density and represent a gaussian distribution centered at the anchor value and with standard deviation of *0.5*, same as the chosen kernel function. Density value as a feature can therefore be misleading in absence of anchor count as a contextual attribute since a single-anchor sample will tend to have higher density, even when compared with multiple-anchor sample with anchors occurring in a close neighborhood. This is demonstrated in Figure 4, where we see that the distribution of kernel density scores shifts lower as number of anchors used for estimation increase. To overcome this, we always use paired features – density estimates and number of anchors in downstream classification models. This layer is used as a feature engineering layer to prepare distance-based transformations, $D$ using the markup-based features, $M$.

### 3.4 Detector Layer

The detector layer sits downstream to the KDE feature generation layer and comprises of three classification models, each with the objective of detecting aberrant values of individual anchors. The reason to have individual classification models rather than having one uber-model for detection is twofold – (1) *coverage* – given that all items do not have all anchor prices available and an uber-model that leverages all anchor price based features can lead to lower decision coverage and (2) *problem isolation* – it can be challenging to pinpoint one or more aberrant anchors as an uber-model can indicate sample-level anomaly but is not concretely able to highlight either feature or anchor-level issues.

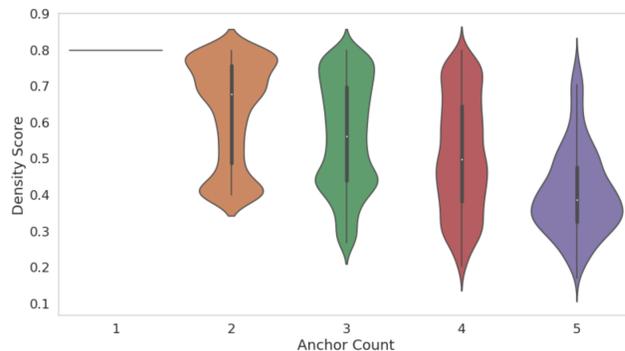

**Figure 4. Distribution of fitted kernel density scores with different anchor price counts using normal samples. As the number of anchor prices increases, average kernel density scores across samples tend to drop. Using kernel density scores as a feature in detector models without context (anchor price count) may lead to higher misclassification rate.**

Additionally, the detector layer has to be highly modular to provide flexibility for additional anchor prices in future – for instance, if a new stream of an additional competitor's price becomes available as an anchor price. With anchor-level detector models, we can achieve a flexible and scalable design with a possibility of enhancing individual anchor-price detectors without disrupting the other detectors or the entire system. At this time, we focus on three anchor prices which have demonstrated higher anomaly ratio past escalations.

Common baseline statistical approaches for outlier detection like distance from center of distribution or bounds based on interquartile range are widely used in literature, however for our use-case did not perform well due to complex nature of anomalies. Prices often fluctuate and high variance alone is not sufficient for capturing anomalies. Relying on statistical approaches results in high false positive rate. We focus our work on classification models which tend to be fast at inference-time while being able to achieve complex optimization objectives to separate anomalies from normal anchor price updates. For training binary classification models which belong to the domain of supervised learning, labeled data is required. Traditionally, we have been working with an internal operations team to review and label potentially anomalous item prices from other pricing anomaly detection models [5]. These human-labeled samples are then used for monitoring performance and retraining the legacy models. For our use-case, leveraging human labels was not the most suitable approach due to the following reasons - (1) it takes significant effort upfront to set up crowdsourced labeling process, (2) human labels in pricing context tend to be less reliable than other cognitive tasks like image or text, and (3) developing the same workflow for each anchor-level model is expensive, requires stakeholder buy-in and can have a long tail before data is available for training models.

We therefore explored other avenues for generating labeled data. Some alternatives to human supervision include (i) distant supervision which involves obtaining noisy labels using external knowledge bases, (ii) weak supervision which leverages labeling functions to encode domain heuristics to build noisy labels and eventually denoising them and (iii) leveraging transfer learning to infer labels from prior models trained for similar task. Considering the absence of prior models for transfer learning task, and existence of public databases related to pricing information, we focused on weak supervision to generate labeled data by encoding weak heuristics as labeling functions.

---

**Labeling function:** a heuristic for labeling competitor prices which are too far from AUR as anomalous

1. **function** high_distance_from_aur (item_data, thresh);
2.     Inputs:
3.                item_data
4.                threshold of distance between cmp price and aur
5.     Output:
6.                items with difference absolute between cmp price and aur greater than thresh
7.     positive_tag_items = set()
8.     **for** (X, y) in (item_data, item_ids) **do**
9.         **if** abs(X.cmp_price − X.aur) > thresh **then**
10.                positive_tag_items.add(y)
11.         **end if**
12.     **end for**
13.     **return** positive_tag_items
14. **end function**

---

We leverage similar but distinct labeling functions for each of the individual detector models. An example of a labeling function is provided, in context of labeling competitor price anchor. **Average Unit Retail** (AUR) is defined as the average order-based price in a time interval. It is used to benchmark reasonable price for items since receiving orders at given price points indicates the prices are reasonable on average. When competitor price is within a reasonable threshold around AUR, it is considered normal, else it is labeled as anomalous.

Some labeling functions were designed exclusively for subsets of data that had dense anchor coverage and naturally, this subset is biased when compared with the population of production price events. To ensure consistent distribution of training set with the data observed in production environment, we perform **anchor masking**. We nullify anchor prices at random to achieve same level of anchor sparsity in the train set as we see in real-world. We then recalculate the set of distance-based input features, $D$ using masked anchors and train detector models.

## 3.5 Aggregator Layer

The aggregator layer sits after the detector layer and aims to use the outlier-cleansed anchor array $C$ to build an optimal anchor. While the detector layer leverages numeric features like historically observed anchor price values $H$ and anchor densities using KDE, the aim of aggregator is to leverage independent information about an item. To that purpose, we introduce additional contextual information in the aggregator layer. Our observation is that item features can indicate which anchor price can be most reliable as an estimate of reasonable price without the prior knowledge of the observed anchor price value. For instance, quality of competitor prices relies on the quality of the match between Walmart item and competitor item listing. If there is a mismatch, competitor price can be unreliable as an estimate for Walmart item's reasonable price. This can be summarized via a hypothetical example of Books being more likely to have a reliable match than *T-shirts* due to presence of hard global identifiers like ISBN number for *Books*. This idea can be generalized to anchor prices other than competitor prices.

Therefore, a key assumption of aggregator is that under different circumstances, different anchors tend to be more reliable. The objective is to assign a weighting scheme to each element in $C$ (i.e., each anomaly-cleansed anchor) so that the weighted mean output is a reliable estimate of reasonable item price. We choose to take weighted mean rather than selecting the maximum probability class to further hedge against incorrect classification by aggregator model.

To obtain reasonable weights, we framed the optimization problem using a multiclass classification approach. For this optimization, we leveraged a subset of items where AUR was available. As we indicated previously, AUR is considered as a reliable estimate of reasonable price. It is however unavailable for >90% of the catalog because of absence of past orders. For preparing the target value $y_{agg}$, we first identified an anchor (which is an element from the real-time anchor vector, $A$) that is closest to the AUR for each item. In cases where there were ties, we broke ties using random assignment. For instance, if online price and store price are both closest to AUR, we pick one of them randomly.

$$y_{agg} = argmin_{A_i}(\mid A_i - AUR \mid) \tag{4}$$

We engineered various features for modeling using the array of competing marketplace offer prices, $O$. Some examples of transformations include minimum, maximum, range and coefficient of variation. We trained various classification models and report on their performance in the Results section. We first estimate the probability of competitor price and of Walmart store prices to be the ideal anchor price, in absence of knowledge of class values (where ideal implies being closest to the reasonable price for the item). We then use these probabilities as weights and calculate a weighted sum of anchors to build an estimate of item's reasonable price. This approach ensures interpretability for merchants as they troubleshoot any escalations where anchor price is unreasonable.

## 3.6 Evaluation Metrics

As we set the system as an ensemble of multiple layers with potentially multiple models. Each layer in the system adds overhead in terms of inference latency, and the tradeoff between accuracy and performance must be clearly assessed. In addition to the standard classification metrics like precision, recall and $F_1$ score, we also consider two interpretable metrics for performance assessment. First, we consider Median Absolute Percentage Error ($M_eAPE$) as the median of absolute percentage difference in predicted anchor from AUR.

$$e_i = \frac{|y_i - AUR_i|}{AUR_i} \quad (5)$$

$$M_eAPE = Median(e_i) \quad (6)$$

Additionally, we label an item as having a *precise anchor* if its Absolute Percentage Error, $e_i$, is within a threshold $t$ of true AUR value. We take the percentage of items that have *precise anchors* as the second metric of interest. It is defined as Precise Anchor Coverage (PAC).

$$p_i = \begin{cases} 1 & if\ e_i < t \\ 0 & otherwise \end{cases} \quad (7)$$

$$PAC = \frac{1}{n}\sum_{i=1}^{n} p_i \quad (8)$$

$M_eAPE$ is an unbounded positive value, and values closer to zero indicate better performance. PAC is bounded between [0,1] and a higher value indicates better results.

## 4. EXPERIMENTS

We cover the experimental setup and results in this section. First, we start by introducing data aggregation and preprocessing methodology. We report the performance of the individual detector and aggregator models during development. Then we demonstrate the baseline performance and impact on incrementally adding components of our system.

## 4.1 Data and Preprocessing

We first prepare a core dataset and perform multiple experiments using different subsets of the dataset. All item-level price changes in the recent six-month window are initially considered. At random, we sample thirty dates from this period and select all observed price events for these dates. We further randomly sample a single price-update event per item. As labeled data is not readily available, target variable preparation for detector and aggregator models is summarized below. Additionally, a summary of sample sizes used for model training is shown in Table 2.

- **Detector Data –** We generate weakly supervised labels using heuristics and labeling functions.

- **Aggregator Data –** Labels are generated based on the anchor that is closest to the observed AUR. In case of ties, one of the multiple anchors that is closest to observed AUR is selected at random.

**Table 2. Sample sizes for development.**

| Model | Train Set | | Test Set | |
|---|---|---|---|---|
| | $N_L$ | $A_L$ | $N_L$ | $A_L$ |
| Detector - Competitor Price | 61,018 | 371 | 15,244 | 104 |
| Detector - Historical Anchor | 62,107 | 166 | 15,538 | 31 |
| Detector - Marketplace Price Anchor | 62,104 | 573 | 15,534 | 127 |
| Aggregator** | 79,962 | | 19,991 | |
| End-to-end System Performance | - | - | 1,066,339 | 224,984 |

\* $N_L$ denotes normal instances and $A_L$ denotes anomalous instances.
\*\* The concept of normal or anomalous does not exist for aggregator.

## 4.2 Performance Evaluation of Individual Models

We use $F_1$ score as the primary metric for classification performance and report on the three Detector model scores in Table 5. The results displayed are only for the best hyperparameter configurations for each model, tuned using 5-fold cross-validation. The performance is reported using the best model configuration on the test set. We observe that Random Forest models outperform other methods for two detector use-cases and XGBoost performs best for one of the three anchors.

We perform similar exercise for the classification model used in aggregator layer and report on multiple classification metrics including precision, recall, $F_1$ score and AUC (area under the precision-recall curve) in Table 6. Finally, while ensemble models tend to perform better than base Decision Trees, the performance of Decision Trees is not significantly worse for detector models' use-case and offer significantly lower latency. As the system grows and additional anchors are added in future, detector models will need to be kept lean unless there is a significant accuracy improvement on using costlier models. We therefore decide to implement Decision Trees for the detector layer models. For Aggregator layer, since we need a single consolidated model, we decided the system can afford to a have more expensive model and therefore implement Random Forest. Table 7 summarizes the performance of the models selected for implementation.

## 4.3 Performance using Different Components

We highlight the performance on using different combinations of the system components and compare it with the baseline (a simple arithmetic mean of all available anchors) and anchors from the existing rule-based system. We used a sample of ~1.3MN items and narrowed down to a subset of 128K items to show the system performance upon using five different configurations. We only report performance on the set where at least one of the detector models flagged an event as anomaly as these items were more likely to be anomalous and need to be addressed by the system. Reporting on the entire set of 1.3MN will result in the metrics being driven by a large set of non-anomalous items, which are of less interest for the system. The results are summarized in Table 3. We start by

**Table 3. Performance summary under different components on item set where detector flagged an anomalous anchor.**

| Configuration | Precision | Recall | $F_1$ score | $M_eAPE$ | PAC | Latency Increase (m.s.) |
|---|---|---|---|---|---|---|
| Current Production System | 0.8414 | 0.5111 | 0.6359 | 0.4979 | 0.5758 | – |
| Arithmetic Mean without Detector | **0.9058** | 0.6172 | 0.7342 | 0.2847 | 0.7888 | 0.73 |
| Aggregator without Detector | 0.8787 | 0.6922 | 0.7744 | 0.2173 | **0.8723** | 10.80 |
| Arithmetic Mean with Detector | 0.8721 | 0.6978 | 0.7753 | 0.2117 | 0.8430 | 2.87 |
| *MoatPlus* (Aggregator with Detector) | 0.8622 | **0.7202** | **0.7848** | **0.1911** | 0.8496 | 12.42 |

**Table 4. Performance summary on different subsets of production data.**

| Subset Definition | $F_1$ score | | | $M_eAPE$ | | | PAC | | |
|---|---|---|---|---|---|---|---|---|---|
| | $E$ | $B$ | $M_p$ | $E$ | $B$ | $M_p$ | $E$ | $B$ | $M_p$ |
| All Items | 0.7208 | 0.8035 | **0.8157** | 0.2710 | 0.1275 | 0.1089 | 0.8246 | 0.9355 | **0.9423** |
| Items Where Detectors Detected an Anomaly | 0.6359 | 0.7341 | **0.7848** | 0.4979 | 0.2847 | **0.1911** | 0.5758 | 0.7888 | 0.8496 |
| Items With At Least One Anomalous Anchor Price | 0.4819 | 0.5969 | **0.6542** | 0.7888 | 0.4568 | **0.3333** | 0.2479 | 0.6775 | 0.7142 |
| Items Where Existing Anchor Price is Non-Precise | 0.4303 | 0.6038 | **0.6717** | 1.1331 | 0.4638 | **0.3013** | - | 0.6661 | 0.7513 |

*$E$ denotes existing system, $B$ denotes baseline and $M_p$ denotes *MoatPlus*.

**Table 5. $F_1$ scores of various supervised anomaly detection models for Detector layer.**

| Model | Detector 1 Competitor Price | Detector 2 Historical Anchor | Detector 3 Marketplace Price |
|---|---|---|---|
| Decision Tree | 0.9231 | 0.8214 | 0.9112 |
| Random Forest | **0.9854** | 0.8667 | **0.9313** |
| LightGBM | 0.9655 | 0.7586 | 0.9084 |
| XGBoost | 0.9804 | **0.8966** | 0.9266 |

**Table 6. Aggregator model experimental results.**

| Model | Precision | Recall | $F_1$ score |
|---|---|---|---|
| Decision Tree | 0.3509 | 0.3527 | 0.3512 |
| Random Forest | 0.4162 | 0.3604 | **0.3734** |
| LightGBM | 0.4335 | 0.3450 | 0.3495 |
| XGBoost | 0.4334 | 0.3391 | 0.3431 |

**Table 7. Performance summary for the selected models.**

| Model | Precision | Recall | $F_1$ score | AUC |
|---|---|---|---|---|
| Detector - Competitor Price | 0.8654 | 0.9890 | 0.9231 | 0.9277 |
| Detector - Historical Anchor | 0.7419 | 0.9200 | 0.8214 | 0.8312 |
| Detector - Marketplace Price | 0.9291 | 0.8939 | 0.9112 | 0.9118 |
| Aggregator | 0.4162 | 0.3604 | 0.3734 | 0.4000 |

reporting the current rule-based system performance – while the precision of 0.8414 is close to the precision of other configurations, recall of 0.5111 is relatively low, leading to a low $F_1$ score. $M_eAPE$ is relatively high and $PAC$ is fairly low, indicating that majority of the items under current system have suboptimal anchors, i.e. anchors are distant from reasonable item prices. We observe a significant lift in recall upon using the *MoatPlus* anchor, which leverages both Detector and Aggregator layers. We also observe minimum *MAPE* when using *MoatPlus* anchor which indicates that on average, *MoatPlus* anchors are closest to reasonable price when comparing with other configurations. *MoatPlus* system adds 12.42 m.s. of latency per prediction, which is under the acceptable range for the marketplace pricing pipeline.

### 4.4 Different Subsets of Production Data

We run *MoatPlus* on different segments of data to generate an optimal anchor and report on the performance in Table 4. For this comparison, we report on $F_1$ score, $M_eAPE$ and $PAC$. We observe a substantial improvement from the existing system, $E$ using either baseline, $B$ or *MoatPlus* -based optimal-anchor, $M_p$. However, we observe only marginal improvements in the metrics from $B$ to $M_p$ when considering the entire set of 1.3MN items. This makes sense because majority of the items have non-anomalous anchors and therefore the impact on overall catalog is minimal. With rare-event detection however, we are more interested in the minority cases that are either anomalous or detected as anomalous by a system.

We therefore consider three "high-vulnerability" subsets of items that are more likely to generate egregious ceiling prices. First, we consider items where at-least one detector model detected an anomalous anchor price. Next, we consider a subset that contains at least one anomalous anchor price when compared with AUR – the anchor price used in production may or may not be anomalous. Finally, we consider the subset of items where the existing production anchor price is non-precise, i.e. $p_i = 0$ in equation (7). In all the three high-vulnerability subsets, we observe a higher lift in all metrics from $B$ to $M_p$.

We conclude that *MoatPlus* is able to detect anomalous anchor prices and build a more reasonable estimate of reasonable item price than the existing system. PAC observed a lift of 12% for the overall set of items. Moreover, *MoatPlus* based anchors significantly outperform the baseline and existing system in scenarios where items are vulnerable, demonstrating lifts of up to 46.63% in cases where at-least one anomalous anchor price was present.

## 4.5 Deployment and Future Work

As we operate in a high volume, high velocity production environment, and given that incorrect anchor prices can cause high-sales items to get incorrectly unpublished, we make the following considerations for deployment.

**Auditing**. Anchor price values and other features may change very frequently and in case of misclassification related escalations from stakeholders, it is imperative to log information to be able to perform root causes analyses. Additionally, auditing necessary information can aid future analyses to identify strengths, areas of improvement and potential model drift. We audit input features and system outputs in a database at inference time.

**Monitoring**. Being in initial stages of deployment, we aim to monitor the system performance over a broad and representative set of items with diverse attributes. We build a manual annotation pipeline that uses diversity sampling to select up to fifty unpublished offers per day for review by an internal operations team. We focus on measuring precision of the system since misclassifications can lead to incorrectly unpublishing a reasonably priced offer and can lead to lost revenue. Additionally, we label individual anchor prices as anomalous or normal to supplement existing noisy labels for retraining future detector models.

**Incremental deployment**. We initially deploy the system on a relatively small subset of items and monitor performance. We gradually onboard additional items on the *MoatPlus* system in phases with careful monitoring.

**Retraining**. There are two situations which require retraining of some or all of the four models in the system – (a) addition of new anchor price and (b) model drift. When a new anchor price is added to the system, a corresponding detector model is trained using weak supervision based labels, and the aggregator model is retrained with a new anchor as an additional class.

In the next phase, we plan to expand them system by (a) including additional anchor prices in the system like new competitor price streams, (b) enriching the existing detector models using manually annotated data to capture any additional types of anomalies that the heuristics based weak supervision may not able to capture, and (c) consolidation and automation of retraining for the underlying models.

## 5. CONCLUSION

We developed a high price anomaly detection system for a growing marketplace platform. We leverage three highly modular components to engineer features, detect feature-level anomalies and build a reliable estimate of reasonable item price. We used kernel density estimation for assigning higher scores to anchor prices that occur in dense neighborhoods and use corresponding anchor price density score as one of the features for detecting aberrant anchor prices. We train multiple supervised detector models that use the density estimate of corresponding anchor price along with historical anchor price based statistical features. We use heuristics based weak supervision for label generation. Finally, we build aggregator layer containing a multiclass classification model that used contextual information to predict the probability of each anchor being closest to reasonable item price. The ensemble helped improve precise anchor coverage in highly vulnerable items by up to 46.6% when compared to existing system and by up to 8.5% when compared with the baseline, introducing only 12ms incremental latency on average.

## 6. ACKNOWLEDGEMENTS

We would like to acknowledge the exceptional support provided by Weihang Ren, Arindam Jain, and Fangjing Fu. Their tireless efforts in data exploration, feature preparation, library review, and performance evaluation have substantially deepened and expanded the scope of our work. Their technical expertise and unwavering commitment have been instrumental in ensuring the overall excellence of our project. Furthermore, we would like to extend our sincere appreciation to our esteemed leadership for their unwavering support and guidance throughout this work. Their visionary leadership, encouragement, and guidance have been instrumental in shaping the direction of this work and facilitating its successful execution. We extend our special thanks to Prakhar Mehrotra, John Hurley, Sanjay Radhakrishnan, Ivan Glushchenko, Manish Joneja, Dan Miller, and Rachel Gabato for their outstanding support. Additionally, our heartfelt gratitude extends to our exceptional pricing data science, engineering, business, and product teams. Their collective expertise and unwavering dedication have significantly accelerated our progress and enriched our findings. We offer our special thanks to Elham Shaabani, Mo Elzayat, Rajendra Pittu, Andrei Parenco, Bisman Preet Singh Sachar, Turk Meesamonyont and Tanvi Agrawal for their exceptional contributions.